\documentclass[runningheads]{llncs}
\usepackage[T1]{fontenc}
\usepackage{graphicx}
\usepackage{times}
\usepackage{soul}
\usepackage{url}
\usepackage{subfigure}
\usepackage{caption}
\usepackage[hidelinks]{hyperref}
\usepackage[utf8]{inputenc}
\usepackage[labelsep=colon,skip=10pt]{caption}
\usepackage{graphicx}
\usepackage{amsmath}
\usepackage{amssymb}
\usepackage{booktabs}
\usepackage{algorithm}
\usepackage{algorithmic}
\usepackage{mathtools}
\usepackage{array}
\usepackage{multirow}
\usepackage{stmaryrd}
\usepackage[misc]{ifsym}
\usepackage{booktabs}

\newcommand{\repthanks}{\textsuperscript{\thefootnote}}

\begin{document}
\title{Scalable Adversarial Online Continual Learning}
%
%
\author{Tanmoy Dam $^{1}$\thanks{equal contribution}, Mahardhika Pratama\Letter $^{2}$ \repthanks, MD Meftahul Ferdaus $^{3}$ \repthanks, Sreenatha Anavatti $^{1}$, Hussein Abbas$^1$}
\authorrunning{T. Dam, M. Pratama et al.}
%
\institute{SEIT, University of New South Wales, Canberra, Australia \\
\email{t.dam@student.adfa.edu.au, \{s.anavatti,h.abbas\}@adfa.edu.au}\\ \and STEM, University of South Australia, Adelaide, Australia \email{dhika.pratama@unisa.edu.au}\and ATMRI, Nanyang Technological University, Singapore\\ \email{mdmeftahul.ferdaus@ntu.edu.sg}}
\toctitle{Lecture Notes in Computer Science}
\toctitle{Tanmoy~Dam}

\maketitle              
\begin{abstract}
Adversarial continual learning is effective for continual learning problems because of the presence of feature alignment process generating task-invariant features having low susceptibility to the catastrophic forgetting problem. Nevertheless, the ACL method imposes considerable complexities because it relies on task-specific networks and discriminators. It also goes through an iterative training process which does not fit for online (one-epoch) continual learning problems. This paper proposes a scalable adversarial continual learning (SCALE) method putting forward a parameter generator transforming common features into task-specific features and a single discriminator in the adversarial game to induce common features. The training process is carried out in meta-learning fashions using a new combination of three loss functions. SCALE outperforms prominent baselines with noticeable margins in both accuracy and execution time.

\end{abstract}
\section{Introduction}
Continual learning (CL) has received significant attention because of its importance in improving existing deep learning algorithms to handle long-term learning problems. Unlike conventional learning problems where a deep model is presented with only a single task at once, a continual learner is exposed to a sequence of different tasks featuring varying characteristics in terms of different distributions or different target classes \cite{Chen2016LifelongML}. Since the goal is to develop a never-ending learning algorithm which must scale well to possibly infinite numbers of tasks, it is impossible to perform retraining processes from scratch when facing new tasks. The CL problem prohibits the excessive use of old data samples and only a small quantity of old data samples can be stored in the memory. 

The CL problem leads to two major research questions. The first question is how to quickly transfer relevant knowledge of old tasks when embracing a new task. The second problem is how to avoid loss of generalization of old tasks when learning a new task. The loss of generalization power of old tasks when learning a new task is known as a catastrophic forgetting problem \cite{Chen2016LifelongML,Parisi2019ContinualLL} where learning new tasks catastrophically overwrites important parameters of old tasks. The continual learner has to accumulate knowledge from streaming tasks and achieves improved intelligence overtime. 

There exists three common approaches for continual learning \cite{Parisi2019ContinualLL}: memory-based approach \cite{LopezPaz2017GradientEM}, structure-based approach \cite{Rusu2016ProgressiveNN}, regularization-based approach \cite{Kirkpatrick2017OvercomingCF}. The regularization-based approach makes use of a regularization term penalizing important parameters of old tasks from changing when learning new tasks. Although this approach is computationally light and easy to implement, this approach does not scale well for a large-scale CL problem because an overlapping region across all tasks are difficult to obtain. The structure-based approach applies a network growing strategy to accommodate new tasks while freezing old parameters to prevent the catastrophic forgetting problem. This approach imposes expensive complexity if the network growing phase is not controlled properly or the structural learning mechanism is often done via computationally expensive architecture search approaches thus being infeasible in the online continual learning setting. The memory-based approach stores a small subset of old samples to be replayed along with new samples to handle the catastrophic forgetting problem. Compared to the former two approaches, this approach usually betters the learning performance. The underlying challenge of this approach is to keep a modest memory size. SCALE is categorized as a memory-based approach here where a tiny episodic memory storing old samples is put forward for experience replay mechanisms. 

The notion of adversarial continual learning (ACL) is proposed in \cite{Ebrahimi2020AdversarialCL}. The main idea is to utilize the adversarial learning strategy \cite{Goodfellow2014GenerativeAN,Ganin2016DomainAdversarialTO} to extract task-aligned features of all tasks deemed less prone to forgetting than task-specific features. It offers disjoint representations between common features and private features to be combined as an input of multi-head classifiers. The main bottleneck of this approach lies in expensive complexities because private features are generated by task-specific networks while common features are crafted by the adversarial game played by task-specific discriminators. In addition, ACL is based on an iterative training mechanism which does not fit for online (single-epoch) continual learning problems. 

This paper proposes scalable adversarial continual learning (SCALE) reducing the complexity of ACL significantly via a parameter generator network and a single discriminator. The parameter generator network produces scaling and shifting parameters converting task-invariant features produced by the adversarial learning mechanism to task-specific features \cite{Perez2018FiLMVR,Pham2021ContextualTN}. Our approach does not need to store task-specific parameters rather the parameter generator network predicts these parameters leading to private features. Production of private features are carried out with two light-weight operations, scaling and shifting. The parameter generator is trained in the meta-learning way using the validation loss of the base network, i.e., feature extractor and classifier. The meta-learning strategy is done by two data partitions: training set and validation set portraying both new and old concepts. The training set updates the base network while the validation set trains the parameter generator. Our approach distinguishes itself from \cite{Pham2021ContextualTN} where the adversarial training approach is adopted to produce task-invariant features and we do not need to construct two different memories as per \cite{Pham2021ContextualTN}. Unlike ACL, the adversarial game is played by a single discriminator without any catastrophic problem while still aligning the features of all tasks well.  

SCALE outperforms prominent baselines with over $1\%$ margins in accuracy and forgetting index while exhibiting significant improvements in execution times. The ablation study, memory analysis and sensitivity analysis further substantiate the advantages of SCALE for the online (one-epoch) continual learning problems. This paper offers four major contributions: 1) a new online continual learning approach, namely SCALE, is proposed; 2) our approach provides a scalable adversarial continual learning approach relying only on a single parameter generator for feature transformations leading to task-specific features and a single discriminator to induce task-invariant features; 3) the training process is done in the meta-learning manner using a new combination of three loss functions: the cross-entropy loss function, the DER++ loss function \cite{Buzzega2020DarkEF} and the adversarial loss function \cite{Ebrahimi2020AdversarialCL}. Although the adversarial loss function already exists in \cite{Ebrahimi2020AdversarialCL}, the adversarial game is done differently here using the concept of BAGAN \cite{Mariani2018BAGANDA} rather than that the gradient reversal strategy \cite{Ebrahimi2020AdversarialCL,Ganin2016DomainAdversarialTO}; 4) All source codes, data and raw numerical results are made available in \url{https://github.com/TanmDL/SCALE} to help further studies.  

 \begin{figure}[ht]
 \centering
 \includegraphics[scale=0.32]{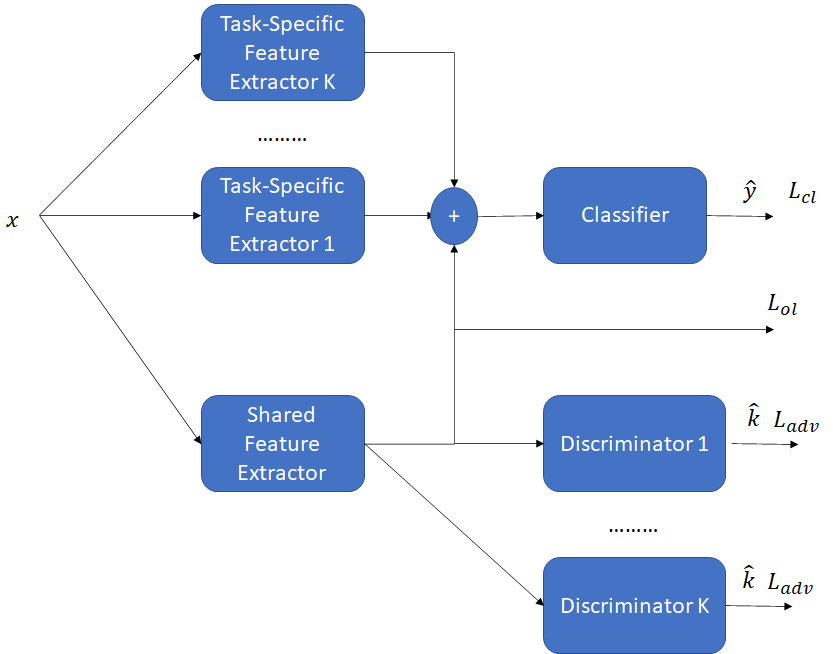}
 \caption{Structure of ACL based on task-specific feature extractors and discriminators.}
 \label{fig_framework}
 \end{figure}

\section{Related Works}
\noindent\textbf{Regularization-based Approach} relies on a penalty term in the loss function preventing important parameters of old tasks from significant deviations. The L2-regularization strategy is combined with the parameter importance matrix indicating the significance of network parameters. Different strategies are proposed to construct the parameter importance matrix: Elastic Weight Consolidation (EWC) makes use of the Fisher Importance Matrix (FIM) \cite{Kirkpatrick2017OvercomingCF}, Synaptic Intelligence (SI) utilizes accumulated gradients \cite{Zenke2017ContinualLT}, Memory Aware Synapses (MAS) adopts unsupervised and online criteria \cite{Aljundi2018MemoryAS}. online EWC (oEWC) puts forward an online version of EWC using Laplace approximation \cite{Schwarz2018ProgressC}. Learning without Forgetting (LWF) utilizes the knowledge distillation (KD) approach to match between current and previous outputs. The regularization strategy is better performed in the neuron level rather than in the synaptic level \cite{Paik2020OvercomingCF} because of the hierarchical nature of the deep neural network. \cite{Mao2021ContinualLV} follows the same principle as \cite{Paik2020OvercomingCF} and goes one step further using the concept of inter-task similarity. Such approach allows a node to be shared across related tasks. Another attempt to improve scalability of regularization-based approaches also exists in \cite{Cha2021CPRCR} where the projection concept is put forward to induce wide local optimum regions. The regularization-based approach heavily depends on the task-IDs and the task-boundaries.

\noindent\textbf{Structure-based Approach} offers different philosophies where new tasks are handled by adding new network components while isolating old components to avoid the catastrophic forgetting problem. The pioneering approach is the progressive neural network (PNN) \cite{Rusu2016ProgressiveNN} where a new network column is integrated when handling a new task. PNN incurs expensive structural complexities when dealing with a long sequence of tasks. \cite{Yoon2018LifelongLW} puts forward a network growing condition based on a loss criterion with the selective retraining strategy. The concept of neural architecture search (NAS) is proposed in \cite{Li2019LearnTG} to select the best action when observing new tasks. Similar approach is designed in \cite{Xu2021AdaptivePC} but with the use of Bayesian optimization approach rather than the NAS concept. These approaches are computationally prohibitive and call for the presence of task IDs and boundaries. \cite{Pratama2021UnsupervisedCL,Ashfahani2022UnsupervisedCL} put forward a data-driven structural learning for unsupervised continual learning problems where hidden clusters, nodes and layers dynamically grow and shrink. The key difference between the two approaches lies in the use of regularization-based approach in \cite{Ashfahani2022UnsupervisedCL} via the Knowledge Distillation (KD) strategy and the use of centroid-based experience replay in \cite{Pratama2021UnsupervisedCL}. The data-driven structural learning strategy does not guarantee optimal actions when dealing with new tasks.

\noindent\textbf{Memory-based Approach} utilizes a tiny memory storing a subset of old data samples. Old samples of the memory are interleaved with current samples for experience replay purposes to cope with the catastrophic forgetting problem. iCaRL exemplifies such approach \cite{Rebuffi2017iCaRLIC} where the KD approach is performed with the nearest exemplar classification strategy. GEM \cite{LopezPaz2017GradientEM} and AGEM \cite{Chaudhry2019EfficientLL} make use of the memory to identify the forgetting cases. HAL \cite{Chaudhry2021UsingHT} proposes the idea of anchor samples maximizing the forgetting metric and constructed in the meta-learning manner. DER \cite{Buzzega2020DarkEF} devises the dark knowledge distillation and successfully achieves improved performances with or without the task IDs. CTN is proposed in \cite{Pham2021ContextualTN} using the feature transformation concept of \cite{Perez2018FiLMVR} and integrates the controller network trained in the meta-learning fashion. ACL \cite{Ebrahimi2020AdversarialCL} is categorized as the memory-based approach where a memory is used to develop the adversarial game. However, ACL imposes considerable complexities because of the use of task-specific feature extractors and discriminators. We offer an alternative approach here where private features are induced by the feature transformation strategy of \cite{Perez2018FiLMVR} and the adversarial game is played by one and only one discriminator. Compared to \cite{Pham2021ContextualTN}, SCALE integrates the adversarial learning strategy to train the shared feature extractor generating common features being robust to the catastrophic forgetting problem and puts forward a new combination of three loss functions.

\section{Problem Formulation}
Continual learning (CL) problem is defined as a learning problem of sequentially arriving tasks $\mathcal{T}_1,\mathcal{T}_2,...,\mathcal{T}_{k}, k\in\{1,...,K\}$ where $K$ denotes the number of tasks unknown in practise. Each task carries triplets $\mathcal{T}_k=\{x_i,y_i,t_i\}_{i=1}^{N_k}$ where $N_k$ stands for a task size. $x_i^k\in\mathcal{X}_k$ denotes an input image while $y_i^k\in\mathcal{Y}_k$, $y_i^k=[l_1,l_2,...,l_m]$ labels a class label and $t_i^k$ stands for a task identifier (ID). The goal of CL problem is to build a continual learner $f_{\phi}(g_{\theta}(.))$ performing well on already seen tasks where $g_{\theta}(.)$ is the feature extractor and $f_{\phi}(.)$ is the classifier. This paper focuses on the online (one-epoch) task-incremental learning and domain-incremental learning problems \cite{Ven2019ThreeSF} where each triplet of any tasks $\{x_i,y_i,t_i\}\backsim\mathcal{T}_k$ is learned only in a single epoch. The task-incremental learning problem features disjoint classes of each task, i.e., Suppose that $L_k$ and $L_{k'}$ stand for label sets of the $k-th$ task and the $k'-th$ task, $\forall k,k' L_k\cap L_{k'}=\emptyset$. The domain-incremental learning problem presents different distributions or domains of each task $P(X,Y)_{k}\neq P(X,Y)_{(k+1)}$ while still retaining the same target classes for each task. That is, a multi-head configuration is applied for the task-incremental learning problem where an independent classifier is created for each task $f_{\phi_k}(.)$. The domain-incremental learning problem is purely handled with a single head configuration $f_{\phi}(.)$. The CL problem prohibits the retraining process from scratch $\frac{1}{K}\sum_{k=1}^{K}\mathcal{L}_{k},\mathcal{L}_{k}\triangleq\mathbb{E}_{(x,y)\backsim\mathcal{D}_k}[l(f_{\phi}(g_{\theta}(x)),y)]$. The learning process is only supported by data samples of the current task $\mathcal{T}_k$ and a tiny memory $\mathcal{M}_{k-1}$ containing old samples of previously seen tasks to overcome the catastrophic forgetting problem.  
\section{Adversarial Continual Learning (ACL)}
Fig. \ref{fig_framework} visualizes the adversarial continual learning method \cite{Ebrahimi2020AdversarialCL} comprising four parts: shared feature extractor, task-specific feature extractors, task-specific discriminator and multi-head classifiers. The shared feature extractor generates task-invariant features while the task-specific feature extractors offer private features of each task. The task-specific discriminator predicts the task's origins while the multi-head classifiers produce final predictions. The training process is governed by three loss functions: the classification loss, the adversarial loss and the orthogonal loss. The classification loss utilizes the cross-entropy loss function affecting the multi-head classifiers, the shared feature extractor and the task-specific feature extractors. The adversarial loss is carried out in the min-max fashion between the shared feature extractor and the task-specific discriminators. The gradient reversal layer is implemented when adjusting the feature extractor thus converting the minimization problem into the maximization problem. That is, the shared feature extractor is trained to fool the task-specific discriminators and eventually generates the task-invariant features. The orthogonal loss ensures clear distinctions between the task-specific features by the shared feature extractor and the private features by the task-specific feature extractors. 

The task-specific discriminator is excluded during the testing phase and the inference phase is performed by feeding concatenated features of the private features and the common features to the multi-head classifiers producing the final outputs. ACL incurs high complexity because of the application of the task-specific feature extractors and the task-specific discriminators. We offer a parameter generator here generating scaling and shifting parameters to perform feature transformation. Hence, the task-specific features are generated with low overheads without loss of generalization, while relying only on a single discriminator to play the adversarial game inducing aligned features. In addition, ACL relies on an iterative training procedure violating the online continual learning requirements whereas SCALE fully runs in the one-epoch setting. 

\begin{figure}[ht]
 \centering
 \includegraphics[scale=0.32]{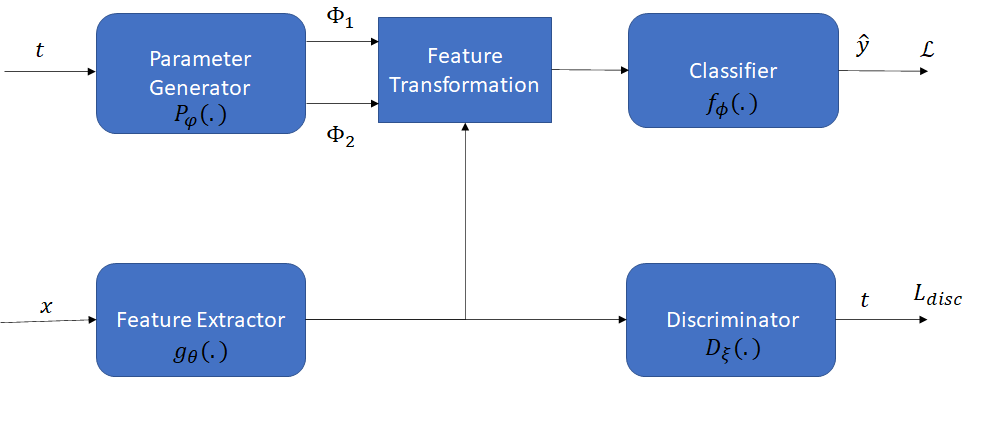}
 \caption{SCALE consists of the parameter generator, the feature extractor, the multi-head classifiers (the single-head classifier in the domain-incremental learning problem) and the discriminator. The parameter generator generates the scaling and shifting coefficients converting the common features into the task-specific features. The task-specific features and the task-invariant features are combined and feed the classifier. The training process is controlled by the classification loss, the DER++ loss and the adversarial loss. The training process of the parameter generator is carried out in the meta-learning fashion minimizing the three loss functions. The single discriminator is updated by playing an adversarial game using the cross entropy loss and the DER++ loss.}
 \label{fig_framework_1}
 \end{figure}

\section{Learning Policy of SCALE}
The learning procedure of SCALE is visualized in Fig \ref{fig_framework_1} and Algorithm 1 where it comprises four blocks: the feature extractor $g_{\theta}(.)$, the parameter generator $P_{\varphi}(.)$, the multi-head classifiers $f_{\phi_k}(.)$ (the single-head classifier in the domain-incremental learning problem) and the single discriminator $D_{\xi}(.)$. The feature extractor extracts the task-invariant features enabled by the adversarial learning mechanism with the discriminator predicting the task IDs. Unlike ACL where task-specific features are produced by task-specific feature extractors, SCALE benefits from the feature transformation strategy with the scaling parameters $\Phi_1$ and the shifting parameters $\Phi_2$ produced by the parameter generator. The scaling and shifting parameters modify the common features into the task-specific features. The classifier receives aggregated features and thus delivers the final predictions. Since the scaling and shifting parameters assure distinct task-specific features of those common features, the orthogonal loss is removed. SCALE replaces the task-specific discriminators in ACL with only a single discriminator.    

\subsection{Feature Transformation}
SCALE does not deploy any task-specific parameters violating the fixed architecture constraint \cite{Pham2021ContextualTN} rather the parameter generator produces the scaling and shifting parameters thereby reducing its complexity significantly. We adopt similar idea of  \cite{Perez2018FiLMVR,Pham2021ContextualTN} where the scaling and shifting parameters creates the task-specific features via the feature transformation procedure as follows:
\begin{equation}
    \Tilde{g}_{\theta}(x)=\frac{\Phi_1}{||\Phi_1||_2}\odot g_{\theta}(x)+\frac{\Phi_2}{||\Phi_2||_2}
\end{equation}
where $\odot$ denotes the element-wise multiplication. $\Phi_1,\Phi_2$ are the scaling and shifting parameters generated by the parameter generator $P_{\varphi}(t)=\{\Phi_1,\Phi_2\}$ taking the task IDs as input features with an embedding layer to produce low-dimensional features. This implies the parameter generator network $\varphi$ to produce the scaling and shifting parameters $\Phi_1$, $\Phi_2$. A residual connection is implemented to linearly combine the shared and private features:
\begin{equation}
    g_{\theta}(x)=\Tilde{g}_{\theta}(x)+g_{\theta}(x)
\end{equation}
We follow the same structure as \cite{Pham2021ContextualTN} where the feature transformation strategy is implemented per layer with one parameter generator per layer. It is implemented for all intermediate layers except for the classifier in the case of multi-layer perceptron network while it is only applied to the last residual layer for convolutional neural network, thus only utilizing a single parameter generator network. A nonlinear activation function $s(.)$ is usually applied before feeding the combined features to the classifier $f_{\phi_k}(s(g_{\theta}(.)))$. 
\subsection{Loss Function}
The loss function of SCALE consists of three components: the cross-entropy (CE) loss function, the dark-experience replay++ (DER++) loss function \cite{Buzzega2020DarkEF}, and the adversarial loss function \cite{Ebrahimi2020AdversarialCL}. The CE loss function focuses on the current task and the previous tasks simultaneously while the DER++ loss function concerns on the past tasks thus distinguishing the second part of the DER++ loss with the CE loss function. The adversarial loss function is designed to align features of all tasks. Suppose that $o=f_{\phi}(g_{\theta}(.))$ stands for the output logits or the pre-softmax responses and $l(.)$ labels the cross-entropy loss function, the loss function of SCALE is expressed:
\begin{equation}\label{overall}
\begin{split}
    \mathcal{L}=\underbrace{\mathbb{E}_{(x,y)\backsim\mathcal{D}_{k}\cup\mathcal{M}_{k-1}}[l(o,y)]}_{L_{CE}}+\underbrace{\mathbb{E}_{(x,y)\backsim\mathcal{M}_{k-1}}[\lambda_{1}||o-h||_2+\lambda_{2}l(o,y)]}_{L_{DER++}}+\\\underbrace{\mathbb{E}_{(x,y)\backsim\mathcal{D}_k\cup\mathcal{M}_{k-1}}[\lambda_{3}l(D_{\xi}(g_{\theta}(x)),t)]}_{L_{adv}}
\end{split}
\end{equation}
where $h=f_{\phi}(g_{\theta}(.))_{k-1}$ is the output logit generated by a previous model, i.e., before seeing the current task. $\lambda_1,\lambda_2,\lambda_3$ are trade-off constants. The second term of $L_{DER++}$, $l(o,y)$, prevents the problem of label shifts ignored when only checking the output logits without the actual ground truth. The three loss functions are vital where the absence of one term is detrimental as shown in our ablation study.  

\subsection{Meta-training Strategy}
The meta-training strategy \cite{schmidhuber:1987:srl,Finn2017ModelAgnosticMF} is implemented here to update the parameter generator $P_{\varphi}(.)$ subject to the performance of the base network $f_{\phi}(g_{\theta}(.))$. This strategy initiates with creation of two data partitions: the training set $\mathcal{T}_{train}^{k}$ and the validation set $\mathcal{T}_{val}^{k}$ where both of them comprise the current data samples and the memory samples $\mathcal{T}^{k}\cup\mathcal{M}_{k-1}$. The meta-learning strategy is formulated as the bi-level optimization problem using the inner loop and the outer loop \cite{Pham2021ContextualTN} as follows:
\begin{equation}\label{bilevel}
    \begin{split}
        Outer: \min_{\varphi}\mathbb{E}_{(x,y)\backsim\mathcal{T}_{val}^{k}}[\mathcal{L}]\\
        Inner: s.t \quad\{\phi^*,\theta^*\}=\arg\min_{\phi,\theta}\mathbb{E}_{(x,y)\backsim\mathcal{T}_{train}^{k}}[\mathcal{L}]
    \end{split}
\end{equation}
where $\mathcal{L}$ denotes the loss function of SCALE as formulated in \eqref{overall}. From \eqref{bilevel}, the parameter generator and the classifier are trained jointly. Because of the absence of ground truth of the scaling and shifting coefficients, our objective is to find the parameters of the parameter generator $\varphi$ that minimizes the validation loss of the base network. This optimization problem is solvable with the stochastic gradient descent (SGD) method where it first tunes the parameters of the classifier in the inner loop:
\begin{equation}\label{inner}
    \{\phi,\theta\}=\{\phi,\theta\}-\alpha\sum_{\mathclap{(x,y)\in\mathcal{T}_{train}^{k}}}\nabla_{\{\phi,\theta\}}[\mathcal{L}]
\end{equation}
where $\alpha$ is the learning rate of the inner loop. Once obtaining updated parameters of the base network, the base network is evaluated on the validation set and results in the validation loss. The validation loss is utilized to update the parameter generator:
\begin{equation}\label{outer}
    \varphi=\varphi-\beta\sum_{\mathclap{(x,y)\in\mathcal{T}_{val}^{k}}}\nabla_{\varphi}[\mathcal{L}]
\end{equation}
where $\beta$ is the learning rate of the outer loop. Every outer loop \eqref{outer} involves the inner loop \label{inner}. Both inner and outer loops might involve few gradient steps as in \cite{Pham2021ContextualTN} but only a single epoch is enforced in SCALE to fit the online continual learning requirements.   
\subsection{Adversarial Training Strategy}
The adversarial training strategy is applied here where it involves the feature extractor $g_{\theta}(.)$ and the discriminator $D_{\xi}(. )$. The goal is to generate task-invariant features, robust against the catastrophic forgetting problem. The discriminator and the feature extractor play a minimax game where the feature extractor is trained to fool the discriminator by generating indistinguishable features while the discriminator is trained to classify the generated features by their task labels \cite{Ebrahimi2020AdversarialCL}. The adversarial loss function $L_{adv}$ is formulated as follows:
\begin{equation}\label{adversarial}
    L_{adv}=\min_{g}\max_{D}\sum_{k=0}^{K}\mathbb{I}_{k=t^k}\log(D_{\xi}(g_{\theta}(x)))
\end{equation}
where the index $k=0$ corresponds to a fake task label associated with a Gaussian noise $\mathcal{N}(\mu,\Sigma)$ while $\mathbb{I}_{k=t^k}$ denotes an indicator function returning $1$ only if $k=t^k$ occurs, i.e, $t^k$ is the task ID of a sample $x$. The feature extractor is trained to minimize \eqref{adversarial} while the discriminator is trained to maximize \eqref{adversarial}. Unlike \cite{Ebrahimi2020AdversarialCL} using the gradient reversal concept in the adversarial game, the concept of BAGAN \cite{Mariani2018BAGANDA} is utilized where the discriminator to trained to associate a data sample to either a fake task label $k=0$ or one of real task labels $k=1,..,K$ having its own output probability or soft label. A generator role is played by the feature extractor. The discriminator is trained with the use of memory as with the base network to prevent the catastrophic forgetting problem where its loss function is formulated:
\begin{equation}\label{l_disc}
    L_{disc}=L_{adv}+L_{DER++}
\end{equation}
where $L_{DER++}$ is defined  as per \eqref{overall} except that the target attribute is the task labels rather than the class labels. Unlike \cite{Pham2021ContextualTN} using two memories, we use a single memory shared across the adversarial training phase and the meta-training phase. 
\begin{algorithm}
\caption{Learning Policy of SCALE}
\begin{algorithmic}
\STATE\textbf{Input}: continual dataset $\mathcal{D}$, learning rates $\mu,\beta,\alpha$, iteration numbers $n_{in}=n_{out}=n_{ad}=1$
\STATE\textbf{Output}: parameters of the base learner $\{\phi,\theta\}$, parameters of the parameter generator $\varphi$, parameters of the discriminator $\xi$
\FOR{$k=1$ to $K$}
\FOR{$n_1=1$ to $n_{out}$} 
\FOR{$n_2=1$ to $n_{in}$}
\STATE Update base learner parameters $\{\theta,\phi\}$ using (4)
\ENDFOR
\STATE Update parameter generator parameters $\varphi$ using \eqref{outer} 
\ENDFOR
\FOR{$n_3=1$ to $n_{ad}$}
\STATE Update discriminator parameters $\xi$ minimizing \eqref{l_disc}
\ENDFOR
\STATE $\mathcal{M}_k=\mathcal{M}_{k-1}\cup B_k$ /*Update memory/*
\ENDFOR
\end{algorithmic}
\end{algorithm}

\begin{table}[]
\centering
\caption{Experimental Details}
\begin{tabular}{c|c|c|c|c|c}
\hline 
Datasets & \#Tasks & \#classes/task & \#training/task & \#testing/task & Dimensions\tabularnewline
\hline 
PMNIST & 23 & 10 & 1000 & 1000 & $1\times28\times28$\tabularnewline
\hline 
SCIFAR-100 & 20 & 5 & 2500 & 500 & $3\times32\times32$\tabularnewline
\hline 
SCIFAR-10 & 5 & 2 & 10000 & 2000 & $3\times32\times32$\tabularnewline
\hline 
SMINIIMAGENET & 20 & 5 & 2400 & 600 & $3\times84\times84$\tabularnewline
\hline 
\end{tabular}
\label{tab:my_label}
\end{table}

\section{Experiments}
The advantage of SCALE is demonstrated here and is compared with recently published baselines. The ablation study, analyzing each learning component, is provided along with the memory analysis studying the SCALE's performances under different memory budgets. All codes, data and raw numerical results are placed in \url{https://github.com/TanmDL/SCALE} to enable further studies. 
\subsection{Datasets}
Four datasets, namely Permutted MNIST (PMNIST), Split CIFAR100 (SCIFAR100), Split CIFAR10 (SCIFAR10) and Split MiniImagenet (SMINIIMAGENET), are put forward to evaluate all consolidated algorithms. The PMNIST features a domain-incremental learning problem with $23$ tasks where each task characterizes different random permutations while the rests focus on the task-incremental learning problem. The SCIFAR100 carries $20$ tasks where each task features $5$ distinct classes. As with the SCIFAR100, the SMINIIMAGENET contains $20$ tasks where each task presents disjoint classes. The SCIFAR10 presents 5 tasks where each task features 2 mutually exclusive classes. Our experimental details are further explained in Table \ref{tab:my_label}.

\subsection{Baselines}
SCALE is compared against five strong baselines: GEM \cite{LopezPaz2017GradientEM}, MER \cite{Riemer2019LearningTL}, ER-Ring \cite{Chaudhry2019OnTE}, MIR \cite{Aljundi2019OnlineCL} and CTN \cite{Pham2021ContextualTN}. The five baselines are recently published and outperform other methods as shown in \cite{Pham2021ContextualTN}. All algorithms are memory-based approaches usually performing better than structure-based approach, regularization-based approach \cite{Pham2021ContextualTN}. No comparison is done against ACL \cite{Ebrahimi2020AdversarialCL} because ACL is not compatible under the online (one-epoch) continual learning problems due to its iterative characteristics. Significant performance deterioration is observed in ACL under the one-epoch setting. All baselines are recently published, thus representing state-of-the art results. All algorithms are executed in the same computer, a laptop with 1 NVIDIA RTX 3080 GPU having 16 GB RAM and 16 cores Intel i-9 processor having 32 GB RAM, to ensure fairness and their source codes are placed in \url{https://github.com/TanmDL/SCALE}. 

\subsection{Implementation Notes}
Source codes of SCALE are built upon \cite{Pham2021ContextualTN,LopezPaz2017GradientEM} and our experiments adopt the same network architectures for each problem to assure fair comparisons. A two hidden layer MLP network with 256 nodes in each layer is applied for PMNIST and a reduced ResNet18 is applied for SCIFAR10/100 and SMINIIMAGENET. The hyper-parameter selection of all consolidated methods is performed using the grid search approach in the first three tasks as with \cite{Chaudhry2019EfficientLL} to comply to the online learning constraint. Hyper-parameters of all consolidated algorithms are detailed in \textbf{the supplemental document}. Numerical results of all consolidated algorithms are produced with the best hyper-parameters. Since the main focus of this paper lies in the online (one-epoch) continual learning, all algorithms run in one epoch. Our experiments are repeated five times using different random seeds and the average results across five runs are reported. Two evaluation metrics, averaged accuracy \cite{LopezPaz2017GradientEM} and forgetting measure \cite{Chaudhry2019EfficientLL} are used to evaluate all consolidated methods. Since all consolidated algorithms make use of a memory, the memory budget is fixed to 50 per tasks. 

\begin{center}
\begin{table}
\caption{Numerical results of consolidated algorithms across the four problems. All methods use the same backbone
network and 50 memory slots per task}\label{maintable}

\begin{centering}
\begin{tabular}{>{\raggedright}p{1.5cm}>{\centering}p{1.8cm}>{\centering}p{1.8cm}>{\centering}p{1.8cm}>{\centering}p{1.8cm}}
\hline 
\multirow{2}{1.5cm}{Method} & \multicolumn{2}{c}{SMINIIMAGENET} & \multicolumn{2}{c}{SCIFAR100}\tabularnewline
\cline{2-5} \cline{3-5} \cline{4-5} \cline{5-5} 
 & ACC($\shortuparrow$) & FM($\shortdownarrow$) & ACC($\shortuparrow$) & FM($\shortdownarrow$)\tabularnewline
\hline 
GEM & 54.50$\pm$0.93 & 7.40$\pm$0.89 & 60.22$\pm$1.07 & 9.04$\pm$0.84\tabularnewline
MER & 53.38$\pm$1.74 & 10.96$\pm$1.57 & 59.48$\pm$1.31 & 10.44$\pm$1.11\tabularnewline
MIR & 54.92$\pm$2.29 & 8.96$\pm$1.68 & 61.26$\pm$0.46 & 9.06$\pm$0.63\tabularnewline
ER & 54.62$\pm$0.80 & 9.50$\pm$1.09 & 60.68$\pm$0.57 & 9.70$\pm$0.97\tabularnewline
CTN  & 63.42$\pm$1.18 & 3.84$\pm$1.26 & 67.62$\pm$0.76 & 6.20$\pm$0.97\tabularnewline
SCALE & \textbf{64.96}$\pm$1.10 & \textbf{2.60}$\pm$0.60 & \textbf{70.24}$\pm$0.76 & \textbf{4.16}$\pm$0.48\tabularnewline
\hline 
\end{tabular}
\par\end{centering}
\centering{}%
\begin{tabular}{>{\raggedright}p{1.5cm}>{\centering}p{1.8cm}>{\centering}p{1.8cm}>{\centering}p{1.8cm}>{\centering}p{1.8cm}}
\hline 
\multirow{2}{1.5cm}{Method} & \multicolumn{2}{c}{PMNIST} & \multicolumn{2}{c}{SCIFAR10}\tabularnewline
\cline{2-5} \cline{3-5} \cline{4-5} \cline{5-5} 
 & ACC($\shortuparrow$) & FM($\shortdownarrow$) & ACC($\shortuparrow$) & FM($\shortdownarrow$)\tabularnewline
\hline 
GEM & 71.10$\pm$0.47 & 10.16$\pm$0.41 &  75.9$\pm$1.3& 12.74$\pm$2.85\tabularnewline
MER & 68.70$\pm$0.35 & 12.04$\pm$0.30 & 79.4$\pm$1.51 & 9.7$\pm$1.07\tabularnewline
MIR & 71.90$\pm$0.49 & 11.74$\pm$0.34 &  79$\pm$1.16& 9.28$\pm$0.91\tabularnewline
ER & 74.76$\pm$0.56 & 9.06$\pm$0.58 &  79.76$\pm$1.26& 8.68$\pm$1.49\tabularnewline
CTN & 78.70$\pm$0.37 & 5.84$\pm$0.36 &  83.38$\pm$0.8& 5.68$\pm$1.43\tabularnewline
SCALE & \textbf{80.70}$\pm$0.46 & \textbf{2.90}$\pm$0.27 &  \textbf{84.9}$\pm$0.91&\textbf{4.46}$\pm$0.4 \tabularnewline
\hline 
\end{tabular}
\end{table}
\par\end{center}

\subsection{Numerical Results}
The advantage of SCALE is demonstrated in Table \ref{maintable} where it outperforms other consolidated algorithms with significant margins. In SMINIIMAGENET, SCALE beats CTN in accuracy with over $1.5\%$ gap and higher than that for other algorithms, i.e., around $10\%$ margin. It also shows the smallest forgetting index compared to other algorithms with over $1\%$ improvement to the second best approach, CTN. The same pattern is observed in the SCIFAR100 where SCALE exceeds CTN by almost $2\%$ improvement in accuracy and shows improved performance in the forgetting index by almost $2\%$ margin. Other algorithms perform poorly compared to SCALE where the accuracy margin is at least over $9\%$. and the forgetting index margin is at least over $5\%$. In PMNIST, SCALE beats its counterparts with at least $2\%$ gap in accuracy while around $2\%$ margin is observed in the forgetting index. SCALE is also the best-performing continual learner in the SCIFAR10 where it produces the highest accuracy with $1.5\%$ difference to CTN and the lowest forgetting index with about $1\%$ gap to CTN. Numerical results of Table \ref{maintable} are produced from five independent runs under different random seeds.  
\begin{figure}
\begin{center}
\subfigure[][]{\includegraphics[scale=0.235]{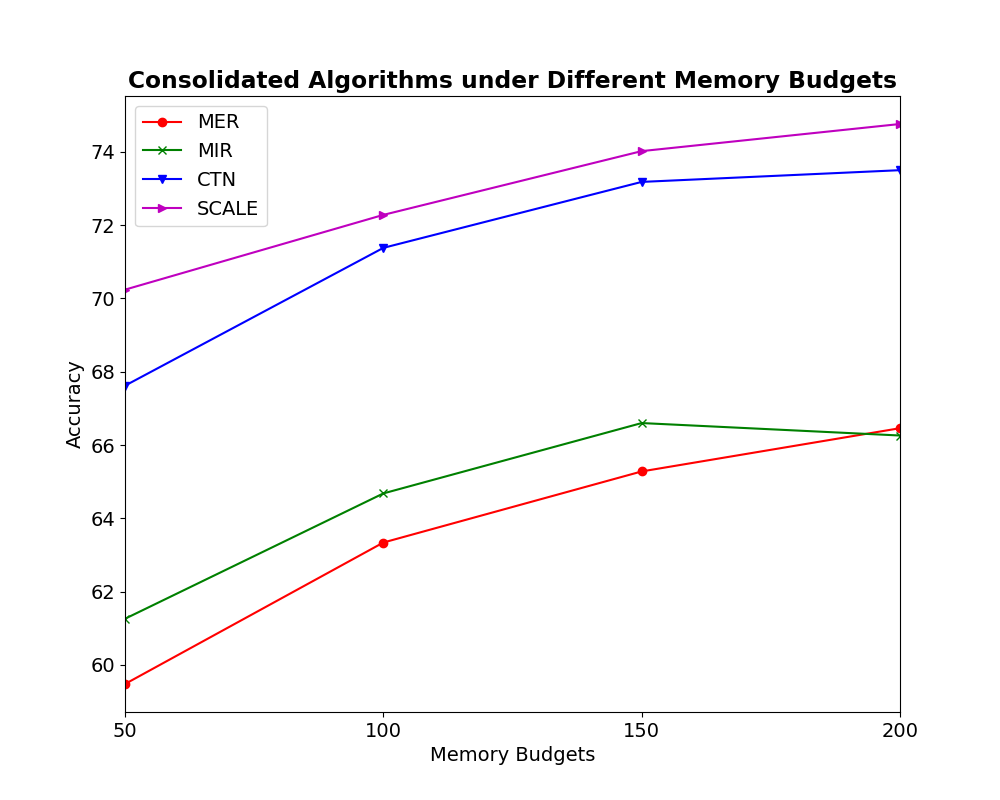}}
\subfigure[][]{\includegraphics[scale=0.235]{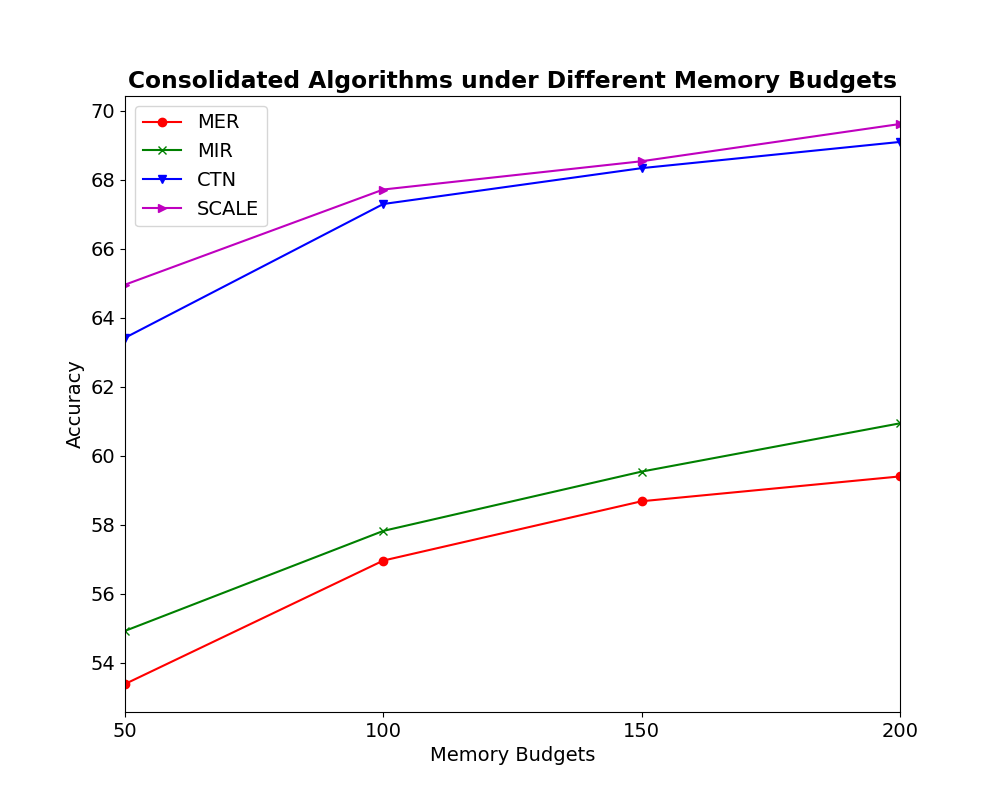}}
\caption{Consolidated Algorithms under Different Memory Budgets in case of (a) SCIFAR100, and (b) SMINIIMAGENET}
\label{fig_memoryanalysis_CIFAR100}
\end{center}
\end{figure}
\subsection{Memory Analysis}
This section discusses the performances of consolidated algorithms, MER, MIR, CTN, SCALE under different memory budgets $|\mathcal{M}_k|=50,100,150,200$ per task. GEM and ER are excluded here because ER performs similarly to MER and MIR while GEM is usually worse than other algorithms. The memory analysis is carried out in the SCIFAR100 and in the SMINIIMAGENET. Our numerical results are visualized in Fig. 2(a) for the SCIFAR100 and in Fig. 2(B) for the SMINIIMAGENET. It is obvious that SCALE remains superior to other algorithms under varying memory budgets in the SCIFAR100 where the gap is at least $1\%$ to CTN as the second best algorithm across all memory configurations. In SMINIIMAGENET problem, SCALE outperforms other algorithms with the most noticeable gap in $|\mathcal{M}_k|=50$ presenting the hardest case. The gap with CTN becomes close when increasing the memory slots per tasks but still favours SCALE. Note that the performances of SCALE and CTN is close to the joint training (upper bound) with increased memory slots in the SMINIIMAGENET, i.e., no room for further performance improvement is possible.   

\begin{center}
\begin{table}
\caption{Ablation Study: Different Learning Configurations of SCALE}\label{ablationstudy}

\begin{centering}
\begin{tabular}{>{\raggedright}p{1.5cm}>{\centering}p{1.8cm}>{\centering}p{1.8cm}>{\centering}p{1.8cm}>{\centering}p{1.8cm}}
\hline 
\multirow{2}{1.5cm}{Method} & \multicolumn{2}{c}{SCIFAR10} & \multicolumn{2}{c}{SCIFAR100}\tabularnewline
\cline{2-5} \cline{3-5} \cline{4-5} \cline{5-5} 
 & ACC($\shortuparrow$) & FM($\shortdownarrow$) & ACC($\shortuparrow$) & FM($\shortdownarrow$)\tabularnewline
\hline 
A & 83.66$\pm$1.42 & 6.26$\pm$1.23 & 68.58$\pm$1.88 & 5.88$\pm$2.06\tabularnewline
B & 76.08$\pm$2.83 & 16.32$\pm$3.57 & 45.32$\pm$2.09 & 27.68$\pm$1.95\tabularnewline
C & 81.68$\pm$1.22 & 7.12$\pm$1.63 & 66.64$\pm$1.71 & 5.76$\pm$0.78\tabularnewline
SCALE & \textbf{84.9}$\pm$0.91 & \textbf{4.46}$\pm$0.40 & \textbf{70.24}$\pm$0.76 & \textbf{4.16}$\pm$0.48\tabularnewline
\hline 
\end{tabular}
\par\end{centering}
\end{table}
\par\end{center}

\subsection{Ablation Study}
This section discusses the advantage of each learning component of SCALE where it is configured into three settings: (A) SCALE with the absence of adversarial learning strategy meaning that the meta-training process is carried out only with the CE loss function and the DER++ loss function while removing any adversarial games; (B) SCALE with the absence of DER++ loss function meaning that the meta-training process is driven by the CE loss function and the adversarial loss function while the adversarial game in \eqref{l_disc} is undertaken without the DER++ loss function; (C) SCALE with the absence of parameter generator network meaning that no task-specific features are generated here due to no feature transformation approaches. Table \ref{ablationstudy} reports our numerical results across two problems: SCIFAR10 and SCIFAR100. 

Configuration (A) leads to drops in accuracy by about $2\%$ and increases in forgetting by about $2\%$ for SCIFAR10 and SCIFAR100. These facts confirm the efficacy of the adversarial learning strategy to boost the learning performances of SCALE. Such strategy allows feature's alignments of all tasks extracting common features, being robust to the catastrophic forgetting problem. Configuration (B) results in major performance degradation in both accuracy and forgetting index across SCIFAR10 and SCIFAR100, i.e., $10\%$ drop in accuracy for SCIFAR10 and $25\%$ drop in accuracy for SCIFAR100; $12\%$ increase in forgetting for SCIFAR10 and $23\%$ increase in forgetting for SCIFAR100. This finding is reasonable because the DER++ loss function is the major component in combatting the catastrophic forgetting problem. Configuration (C) leaves SCALE without any task-specific features, thus causing drops in performances. $3\%$ drop in accuracy is observed for SCIFAR10 while $4\%$ degradation in accuracy is seen for SCIFAR100. The same pattern exists for the forgetting index where $3\%$ increase in forgetting occurs for SCIFAR10 and $1.5\%$ increase in forgetting happens for SCIFAR100. Our finding confirms the advantage of each learning component of SCALE where it contributes positively to the overall performances.

\begin{table}[t!]
\caption{Execution Times of All Consolidated Algorithms across All Problems}
\label{Execution times}
\begin{center}
\scalebox{0.8}{
\begin{tabular}{llccccr}
\toprule
Dataset & Methods & Execution Times \\
\midrule
PMNIST & SCALE & 41.33 \\
& CTN & 52.01 \\
& ER & 26.28 \\
& MIR & \textbf{23.56}\\
& MER & 26.80 \\
& GEM & 28.53 \\
\midrule
SCIFAR10 & SCALE & \textbf{142.1} \\
& CTN & 358.5 \\
& ER & 250.86 \\
& MIR & 242.52 \\
& MER & 257.44 \\
& GEM & 151.96 \\
\midrule
SCIFAR100 & SCALE & \textbf{108.33} \\
& CTN & 321.87 \\
& ER & 211.94 \\
& MIR & 218.24 \\
& MER & 222.39 \\
& GEM & 314.24\\
\midrule
SMINIIMAGENET & SCALE & \textbf{193.73} \\
& CTN & 298.33 \\
& ER & 290.27 \\
& MIR & 254.20 \\
& MER & 261.17 \\
& GEM & 540.46 \\

\bottomrule
\end{tabular}
}

\end{center}
\end{table}

\subsection{Execution Times}
Execution times of all consolidated algorithms are evaluated here because it is an important indicator in the online continual learning problems. Table \ref{Execution times} displays execution times of consolidated algorithms across all problems. The advantage of SCALE is observed in its low running times compared to other algorithms in three of four problems except in the pMNIST. SCALE demonstrates significant improvements by almost $50\%$ speed-up from CTN in realm of execution times because it fully runs in the one-epoch setting whereas CTN undergoes few gradient steps in the inner and outer loops. Note that both SCALE and CTN implement the parameter generator network. This fact also supports the adversarial learning approach of SCALE, absent in CTN, where it imposes negligible computational costs but positive contribution to accuracy and forgetting index as shown in our ablation study. Execution times of SCALE are rather slow in pMNIST problem because the parameter generator is incorporated across all intermediate layers in the MLP network. The execution times significantly improves when using the convolution structure because the parameter generator is only implemented in the last residual block. SCALE only relies on one and only discriminator to produce aligned features while private features are generated via parameter generator networks.  

\begin{table}[]
\centering
\caption{Sensitivity Analysis of Hyper-parameters in SCIFAR100}
\begin{tabular}{c|c|c}
\hline 
Hyper-parameters & ACC ($\shortuparrow$) & FM($\shortdownarrow$) \tabularnewline
\hline 
$\lambda_1,\lambda_2=3,\lambda_3=0.9$ & 69.5$\pm$0.54 & 5.5$\pm$0.44 \tabularnewline
\hline 
$\lambda_1,\lambda_2=3,\lambda_3=0.3$ & 69.34$\pm$1.04 & 5.12$\pm$0.53  \tabularnewline
\hline 
$\lambda_1,\lambda_2=3,\lambda_3=0.09$ & 69.26$\pm$1.22 & 5.2$\pm$1.08 \tabularnewline
\hline 
$\lambda_1,\lambda_2=3,\lambda_3=0.03$ & 69.88$\pm$0.98 & 4.82$\pm$0.55 \tabularnewline
\hline
$\lambda_1,\lambda_2=1,\lambda_3=0.9$ & 69.2$\pm$0.76 & 5.16$\pm$0.85 \tabularnewline
\hline
$\lambda_1,\lambda_2=1,\lambda_3=0.3$ & 69.24$\pm$0.84 & 5.44$\pm$1.19 \tabularnewline
\hline
$\lambda_1,\lambda_2=1,\lambda_3=0.09$ & 69.8$\pm$0.73 & 4.88$\pm$0.90 \tabularnewline
\hline 
$\lambda_1,\lambda_2=1,\lambda_3=0.03$ & \textbf{70.24}$\pm$0.76 & \textbf{4.16}$\pm$0.48 \tabularnewline
\hline 
\end{tabular}
\label{sensitivity}
\end{table}

\section{Sensitivity Analysis}
Sensitivity of different hyper-parameters, $\lambda_1,\lambda_2,\lambda_3$, are analyzed here under the SCIFAR100 where these hyper-parameters control the influence of each loss function \eqref{overall}. Other hyper-parameters are excluded from our sensitivity analysis because they are standard hyper-parameters of deep neural networks where their effects have been well-studied in the literature. Note that the hyper-parameter sensitivity is a major issue in the online learning context because of time and space constraints for reliable hyper-parameter searches. Specifically, we select $\lambda_1,\lambda_2=1$ and $\lambda_1,\lambda_2=3$, while varying $\lambda_3=[0.03,0.09,0.3,0.9]$. Table \ref{sensitivity} reports our numerical results. 

It is observed that SCALE is not sensitive to different settings of hyper-parameters. That is, there does not exist any significant gaps in performances compared to the best hyper-parameters as applied to produce the main results in Table \ref{maintable}, $\lambda_1,\lambda_2=1,\lambda_3=0.03$, the gaps are less than $<1\%$. Once again, this finding confirms the advantage of SCALE for deployments in the online (one-epoch) continual learning problem.  

\section{Conclusion}
This paper presents an online (one-epoch) continual learning approach, scalable adversarial continual learning (SCALE). The innovation of SCALE lies in one and only one discriminator in the adversarial games for the feature alignment process leading to robust common features while making use of the feature transformation concept underpinned by the parameter generator to produce task-specific (private) features. Private features and common features are linearly combined with residual connections where aggregated features feed the classifier for class inferences. The training process takes place in the strictly one-epoch meta-learning fashion based on a new combination of the three loss functions. Rigorous experiments confirm the efficacy of SCALE beating prominent algorithms with noticeable margins ($>1\%$) in accuracy and forgetting index across all four problems. Our memory analysis favours SCALE under different memory budgets while our ablation study demonstrates the advantage of each learning component. In addition, SCALE is faster than other consolidated algorithms in 3 of 4 problems and not sensitive to hyper-parameter selections. Our future work is devoted to continual time-series forecasting problems.  
\subsubsection{Acknowledgements} 
T. Dam acknowledges UIPA scholarship from the UNSW PhD program. M. Pratama acknowledges the UniSA start up grant. 

%
%
%
 \bibliographystyle{splncs04}
 \bibliography{references}
%




\end{document}


\title{Scalable Adversarial Online Continual Learning $Supplementary~Materials$}

\author{Tanmoy Dam $^{1}$\thanks{equal contribution}, Mahardhika Pratama\Letter $^{2}$ \repthanks, MD Meftahul Ferdaus $^{3}$ \repthanks, Sreenatha Anavatti $^{1}$, Hussein Abbas$^1$}

\institute{SEIT, University of New South Wales, Canberra, Australia \\
\email{t.dam@student.adfa.edu.au, \{s.anavatti,h.abbas\}@adfa.edu.au}\\ \and STEM, University of South Australia, Adelaide, Australia \email{dhika.pratama@unisa.edu.au}\and ATMRI, Nanyang Technological University, Singapore\\ \email{mdmeftahul.ferdaus@ntu.edu.sg}}

\authorrunning{T. Dam, M. Pratama et al.}

\toctitle{Lecture Notes in Computer Science}
\toctitle{Tanmoy~Dam}
\maketitle 

This supplementary document is comprised of following two sections.
\begin{itemize}
    \item \textbf{Section 1}: Properties of our proposed SCALE along with baselines with respect to online continual learning setting.
    \item \textbf{Section 2}: hyperparameter configurations for all the methods.
\end{itemize}

\section{Task Specifications}
In this work, task orders are not randomized or optimized. In case of SCIFAR-10, SCIFAR-100, and  SMINIIMAGENET, same order of tasks as in the original datasets are maintained, whereas, it is random (by default) in pMNIST. Main features of our proposed SCALE along with baselines with respect to online continual learning setting is presented in Table 1.

\begin{table}
\caption{Properties of our proposed SCALE along with baselines with respect
to online continual learning setting}
\label{Properties of Algorithms}
\centering{}%
\begin{tabular}{|>{\centering}p{2cm}|>{\centering}p{2cm}|>{\centering}p{2cm}|>{\centering}p{1.3cm}|>{\centering}p{2.7cm}|>{\centering}p{1.5cm}|}
\hline 
Method & Episodic Memory & Task-Specific Parameters & Require Task ID During Inference & Store Historical Params/Grads/Logits (training)  & Store Historical Params (testing)\tabularnewline
\hline 
GEM & $\checked$ & $\times$ & $\checked$ & $\checked$ & $\times$\tabularnewline
\hline 
MER & $\checked$ & $\times$ & $\checked$ & $\checked$ & $\times$\tabularnewline
\hline 
MIR & $\checked$ & $\times$ & $\checked$ & $\times$ & $\times$\tabularnewline
\hline 
ER & $\checked$ & $\times$ & $\checked$ & $\checked$ & $\times$\tabularnewline
\hline 
CTN  & $\checked$ & $\checked$ & $\checked$ & $\checked$ & $\times$\tabularnewline
\hline 
SCALE & $\checked$ & $\checked$ & $\checked$&  $\checked$ &$\times$ \tabularnewline
\hline 
\end{tabular}
\end{table}

\section{Hyperparameter Specifications}
Hyperparameter settings of consolidated methods are presented as follows:

\begin{enumerate}[label*=\arabic*.]
  \item GEM
    \begin{enumerate}[label*=\arabic*.]
    \item Learning Rate: 0.03  (PMNIST, SCIFAR-100 and SCIFAR-10)  and  0.05 ( SMINIIMAGENET)
    \item Number of gradient updates: 1 (all benchmarks)
    \item Margin for QP: 0.5 (all benchmarks)
  \end{enumerate}
  \item MER
  \begin{enumerate}[label*=\arabic*.]
    \item Learning Rate: 0.03  (PMNIST)  and  0.05 (SCIFAR-100, SCIFAR-10 and  SMINIIMAGENET),
    \item Replay batch size: 64 (all benchmarks)
    \item Reptile rate $\beta$: 0.3 (all benchmarks)
    \item Number of gradient updates: 3
  \end{enumerate}
  \item MIR
    \begin{enumerate}[label*=\arabic*.]
    \item Learning Rate: 0.03 (all benchmarks)
    \item Replay batch size: 10 (all benchmarks)
    \item Number of gradient updates: 3
  \end{enumerate}
  
  \item ER
  \begin{enumerate}[label*=\arabic*.]
    \item Learning Rate: 0.03 (all benchmarks)
    \item Replay batch size: 10 (all benchmarks)
    \item Number of gradient updates: 3
  \end{enumerate}
  
    \item CTN
  \begin{enumerate}[label*=\arabic*.]
    \item Inner learning rate: 0.03 (PMNIST) 0.01 (SCIFAR-100, SCIFAR-10 and  SMINIIMAGENET) 
    \item  Outer learning rate: 0.1 (PMNIST) 0.05 (SCIFAR-100, SCIFAR-10 and  SMINIIMAGENET)
    \item Number of inner \& outer updates: 2 (all benchmarks)
    \item Temperature and weight for KL: 5, 100 (all benchmarks)
    \item Replay batch size: 64 (all benchmarks)
    \item Semantic memory percentage: 20\%
  \end{enumerate}
  
      \item SCALE
  \begin{enumerate}[label*=\arabic*.]
    \item Inner learning rate: 0.1 (PMNIST) 0.01 (SCIFAR-100, SCIFAR-10 and  SMINIIMAGENET) 
    \item  Outer learning rate: 0.01 (PMNIST) 0.1 (SCIFAR-100, SCIFAR-10 and  SMINIIMAGENET)
    
    \item  Adversarial learning rate: 0.001 (PMNIST, SCIFAR-100, SCIFAR-10 and  SMINIIMAGENET)
    \item Number of inner \& outer updates: 1 (all benchmarks)
    \item Number of discriminator update: 1 (all benchmarks)
    \item Weights of $\lambda_1,\lambda_2= 1,~\lambda_3= 0.03$ (all benchmarks)
    \item Replay batch size: 64 (all benchmarks)
    
  \end{enumerate}
  
\end{enumerate}

When doing cross-validation across the three validation tasks, grid search is performed to ensure that each hyper-parameter is consistent across all three tasks, which will not be seen during continuous learning..

\begin{itemize}
    \item $\lambda_1,\lambda_2 = \{ 1, 3 \}$
     \item $\lambda_3 = \{ 0.03,0.09, 0.3, 0.9 \}$
     \item $\alpha = \{ 0.001,0.01, 0.1\}$
     \item $\beta = \{ 0.001,0.01, 0.1 \}$
\end{itemize}